\documentclass{article}

\usepackage{amsmath} 
\usepackage[numbers]{natbib}
\usepackage{multirow}
\usepackage{amsmath}
\usepackage{booktabs} 
\usepackage{multirow} 
\usepackage{array}
\usepackage{wrapfig}
\usepackage{microtype}
\usepackage{algpseudocode}
\usepackage{algorithm}
\usepackage{parskip}
\usepackage[export]{adjustbox}
\usepackage{subcaption} 
\usepackage{graphicx}
\usepackage{subcaption} 
\usepackage{threeparttable}
\usepackage{graphicx}
\usepackage[preprint]{neurips_2025}

\usepackage[utf8]{inputenc} 
\usepackage[T1]{fontenc}    
\usepackage{hyperref}       
\usepackage{url}            
\usepackage{booktabs}       
\usepackage{amsfonts}       
\usepackage{nicefrac}       
\usepackage{microtype}      
\usepackage{xcolor}         

\title{Unveiling and Steering Connectome Organization with Interpretable Latent Variables}

\author{%
  Yubin Li$^{1,2}$  Xingyu Liu$^{1}$ Guozhang Chen$^1$\\
  1.National Key Laboratory for Multimedia Information Processing,\\ School of Computer Science, Peking University, Beijing, China\\
  2.School of Computer Science and Engineering, University of Electronic Science and \\Technology of China, Chengdu, Sichuang, China\\
  \texttt{guozhang.chen@pku.edu.cn} \\
}

\begin{document}

\maketitle

\begin{abstract}
The brain's intricate connectome, a blueprint for its function, presents immense complexity, yet it arises from a compact genetic code, hinting at underlying low-dimensional organizational principles. This work bridges connectomics and representation learning to uncover these principles. We propose a framework that combines subgraph extraction from the Drosophila connectome, FlyWire, with a generative model to derive interpretable low-dimensional representations of neural circuitry. Crucially, an explainability module links these latent dimensions to specific structural features, offering insights into their functional relevance. We validate our approach by demonstrating effective graph reconstruction and, significantly, the ability to manipulate these latent codes to controllably generate connectome subgraphs with predefined properties. This research offers a novel tool for understanding brain architecture and a potential avenue for designing bio-inspired artificial neural networks.
\end{abstract}

\section{Introduction}
The principle that structure dictates function is fundamental to understanding complex systems, from biological circuits to artificial intelligence~\cite{chen2022data,ostojic2024computational,liang2024neural-architecture,cai2024neural-architecture2}. In neuroscience, a primary goal is to decipher how the brain's intricate wiring—its connectome—underlies cognition and behavior~\cite{sporns2011, seung2012}. Concurrently, artificial neural networks (ANNs), often drawing inspiration from brain architecture, have achieved significant successes~\cite{gao2025snn,sun2025snn1,guo2022snn2}, yet the principles for their optimal design are still evolving~\cite{++hassabis2017, lecun2015}. The brain remains a key source of inspiration for developing more powerful AI.

However, directly analyzing the brain's complete structural blueprint is a monumental task. The connectome of even a relatively simple organism like the fruit fly, as detailed by datasets like FlyWire~\cite{dorkenwald2022flywire}, involves more than 100,000 neurons. The fly brain contains, in principle, 10,000,000,000 possible connections. Such scale presents significant analytical challenges, while the connection is sparse~\cite{dorkenwald2022flywire}. This immense complexity contrasts with a key biological constraint: the ``genetic bottleneck''~\cite{shuvaev2024pnas}. The genetic code, which carries the instructions for building the brain, can store far less information (approximately 1 GB)~\cite{figureau1987genebottle} than is required to explicitly define every connection in the connectome~\cite{zador2019, marblestone2016}. This disparity strongly implies the existence of a compressed, low-dimensional representation or a set of generative rules encoded in the genome, capable of unfolding into the brain's complex architecture.

Discovering and understanding these low-dimensional representations would be transformative. It could offer deep insights into neural development and the core principles of circuit design. Furthermore, it could provide a means to systematically manipulate these compact codes, paving the way to explore how structural variations impact function and ultimately guide the design of more advanced ANNs. This is the central motivation for our work: to identify interpretable, low-dimensional representations of neural connectivity and use them for analysis and controlled generation. 

To investigate this, we utilize the FlyWire Drosophila connectome, a rich resource for studying whole-brain connectivity. Our aim is to develop a framework that can distill its complex connectivity patterns into compact, understandable latent representations. In this paper, we present a novel framework for interpretable connectome analysis via graph representation learning. Our key contributions are:
\begin{enumerate}
    \item A method combining functionally-guided subgraph sampling with a variational autoencoder (VAE) to derive low-dimensional representations of connectome subgraphs, preserving essential neural connectivity features.
    \item An explainability module using SHAP (SHapley Additive exPlanations) values to link the learned latent dimensions to specific, structurally relevant properties of the neural circuits.
    \item A demonstration of controlled graph generation, where manipulating these interpretable low-dimensional representations enables the targeted synthesis of connectome subgraphs with desired characteristics.
\end{enumerate}

To our knowledge, this integrated approach—from biologically-informed sampling and VAE-based compression to SHAP-driven interpretability and controlled generation for connectome analysis—is a novel endeavor. We validate our method on the FlyWire dataset, showing its effectiveness in graph reconstruction, identifying meaningful patterns, and offering a robust approach for analyzing neural circuit organization.


\section{Problem Definition}
\subsection{Low-dimensional analysis of brain neuron connections}

With advances in neurotechnology, the connectomes of various organisms are being progressively mapped. Understanding the characteristics of neuronal connectivity in the brain holds significant implications for artificial intelligence development. However, even the connectome of simple organisms like the fruit fly comprises hundreds of thousands of neurons and tens of millions of synaptic edges, posing substantial analytical challenges. To address this complexity, we adopt a two-stage approach: first sampling smaller, functionally relevant subgraphs, then compressing them into compact latent vectors. These vectors capture the essential topological and biological features of the subgraphs, enabling efficient comparative analysis and pattern discovery.

\subsection{Variational Autoencoder on Graph}

As the goal is to find latent representations similar to ``genetic bottleneck'', we need a generative model with ``information bottleneck''.
The Variational Autoencoder (VAE) framework provides an effective approach for learning low-dimensional representations of graph-structured data~\cite{kingma2013auto}. Given an input graph $G = (X, A)$ consisting of node features $X \in \mathbb{R}^{N \times d}$ and adjacency matrix $A \in \{0,1\}^{N \times N}$, the model learns to encode the graph into a latent space through a probabilistic mapping:

\begin{equation}
q_\phi(L|G) = \mathcal{N}(\mu_\phi(G), \sigma^2_\phi(G)),
\end{equation}

where $\mu_\phi$ and $\sigma_\phi$ are parameterized by graph neural networks that aggregate structural information. The decoder reconstructs the graph from latent representations through:

\begin{equation}
p_\theta(G|L) = p_\theta(A|L)\prod_{v\in V}p_\theta(X_v|L).
\end{equation}

The model is trained by maximizing the evidence lower bound (ELBO):

\begin{equation}
\mathcal{L} = \mathbb{E}_{q_\phi(L|G)}[\log p_\theta(G|L)] - D_{\text{KL}}(q_\phi(L|G)\|\mathcal{N}(0,I)).
\end{equation}

This objective function ensures both accurate reconstruction of graph properties and well-regularized latent representations, making it particularly suitable for analyzing complex connectome data.
The complete derivation of this objective function is provided in Appendix~\ref{app:vae}.

\section{Methods}
\subsection{Dataset and Sampling Method}
\label{sec:dataset}

The FlyWire dataset\footnote{The dataset is publicly available at \url{https://codex.flywire.ai/api/download}.} provides a complete spatially-embedded connectome of the fruit fly brain~\cite{dorkenwald2022flywire}, containing 3D coordinates for all neurons, neurotransmitter-based classification labels, and directed synaptic connectivity matrices. We use the neuronal connectivity organization from the visual part of the fruit fly brain. 

To analyze this complex network, we employ a adaptive cylindrical sampling strategy along the axis of the visual stream to extract functionally relevant subgraphs. The sampling initiates by randomly selecting a center point in the $XZ$-plane and generating an infinite cylinder aligned with the $Y$-axis to maintain the complete structure of the fruit fly's visual flow. The cylinder radius is dynamically adjusted through binary search to ensure each sample contains 81--100 neurons, balancing biological relevance with computational tractability.

Sampled subgraphs are standardized to 100 nodes through padding with isolated non-neuronal cells, which we classify into a separate category to ensure our model learns meaningful representations. This approach captures columnar functional units characteristic of neural circuitry while accommodating spatial variations in neuronal density. The final dataset comprises 500 such subgraphs, each represented as a directed graph with 100 nodes and 5-dimensional feature vectors encoding four neurotransmitter types(GABA, GLUT, ACH, SER) and one non-neuronal category.

\subsection{Model Structure}
\label{sec:model}
\begin{figure}[h]

	     \begin{subfigure}[b]{1\textwidth}
        \includegraphics[width=\linewidth]{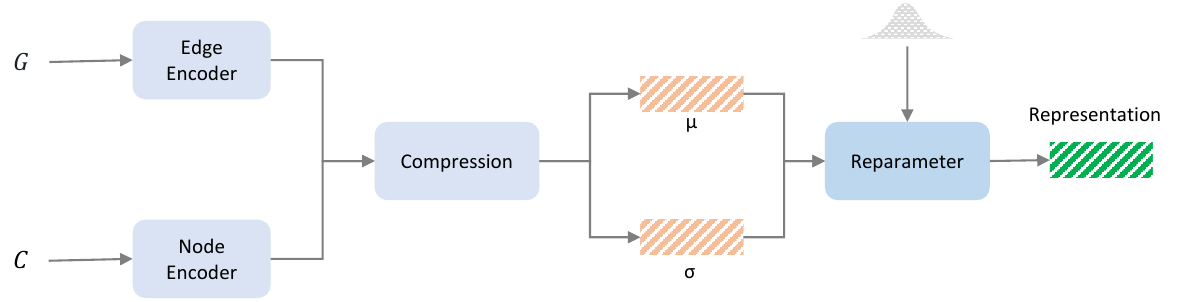}
        \caption{Encoder}
        \label{fig:encoder}
    \end{subfigure}
    \hfill
    \begin{subfigure}[b]{1\textwidth}
        \includegraphics[width=\linewidth]{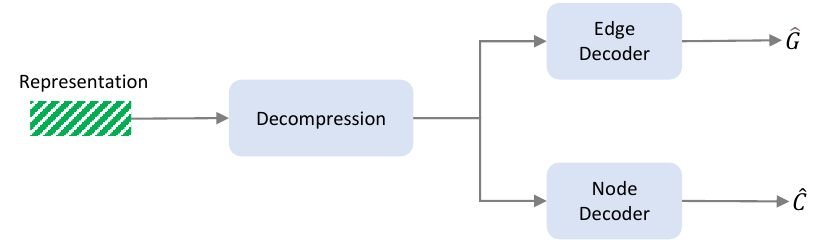}
        \caption{Decoder}
        \label{fig:decoder}
    \end{subfigure}
    \hfill
    \begin{subfigure}[b]{1\textwidth}
        \includegraphics[width=\linewidth]{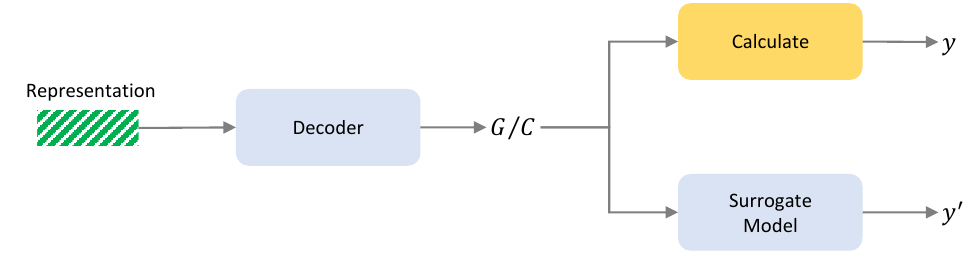}
        \caption{Surrogate model}
        \label{fig:surrogate}
    \end{subfigure}
	
	\caption{Model architecture: (a) shows the encoder module which takes the graph adjacency matrix $G$ and node categories $C$ as input and outputs a first-order low-dimensional representation vector; (b) presents the decoder that reconstructs the graph $(\hat{G}, \hat{C})$ from the latent representation;(c) illustrates the surrogate model designed for SHAP analysis, which takes graph $G$ as input and outputs the approximated feature $y'$.}
	\label{fig:1}
\end{figure}

As shown in Figure~\ref{fig:1}, the proposed model processes an input graph $G=(\mathbf{C},\mathbf{A})$ with categorical node labels $\mathbf{C}\in\mathbb{R}^{N\times1}$ and adjacency matrix $\mathbf{A}\in\mathbb{R}^{N\times N}$, generating a reconstructed graph $G^{\prime}=(\mathbf{C}^{\prime},\mathbf{A}^{\prime})$ through sequential encoding and decoding operations. The edge encoder employs a multi-head Graph Attention Network (GAT)~\cite{velivckovic2017graph} to process the adjacency matrix, where we configure $\mathbf{A}$ to simultaneously serve as both topology input and feature matrix ($\mathbf{H}=\mathbf{A}$) to maximally preserve the connectivity information of the original graph. The architecture implements $K$ parallel attention heads whose outputs are concatenated and linearly projected: $\mathbf{H}^{\prime}=\text{Linear}\big(\big\lVert_{k=1}^{K}\text{GAT}_k(\mathbf{A},\mathbf{H})\big)$, combining multi-scale structural information into edge embeddings.Detailed GAT mechanics are provided in Appendix~\ref{app:gat}.

Node features are encoded through one-hot transformation of categorical labels $\mathbf{C}$, generating $X_{i,j}=\mathbb{I}(C_i=j)\in\mathbb{R}^{N\times F_X}$, where $F_X$ denotes the number of distinct classes. The edge embeddings $\mathbf{H}^{\prime}$ and node features $\mathbf{X}$ are then concatenated into a joint representation $\mathbf{T}=(\mathbf{H}^{\prime}\|\mathbf{X})\in\mathbb{R}^{N\times(F_{H^{\prime}}+F_{X})}$, which is flattened and processed through an MLP with LeakyReLU activation to produce intermediate features $\mathbf{L}^{\prime}=\text{LeakyReLU}(\text{MLP}(\text{flatten}(\mathbf{T}))$.

Separate linear layers generate the latent distribution parameters $\mu=\mathbf{fc}_\mu(\mathbf{L}^{\prime})$ and $\sigma=\mathbf{fc}_\sigma(\mathbf{L}^{\prime})$, enabling probabilistic sampling via the reparameterization trick $\mathbf{z}=\mu+\epsilon\odot\sigma$ where $\epsilon\sim\mathcal{N}$. This stochastic sampling preserves differentiability during backpropagation while capturing uncertainty in the latent space.

For graph reconstruction, the latent code $\mathbf{z}$ is first projected through $\hat{\mathbf{L}}=\text{MLP}(\mathbf{z})$ and reshaped to match the encoder's output structure $\hat{\mathbf{T}}\in\mathbb{R}^{N\times(F_{H^{\prime}}+F_{X})}$. The tensor is split into edge features $\hat{\mathbf{H}}$ and node features $\mathbf{X}^{\prime}$, which are processed by specialized decoders. The edge decoder generates a probabilistic adjacency matrix $\mathbf{A}^{\prime}=\sigma(\text{MLP}(\hat{\mathbf{H}}))$ through sigmoid activation, with final discrete values obtained via thresholding $\mathbf{A}_{\text{pred}}[i,j]=\mathbb{I}(\mathbf{A}^{\prime}[i,j]\geq \kappa)$. Node categories are reconstructed through dimension-wise argmax operations $\hat{C}_i=\arg\max_{1\leq j\leq F_X}\mathbf{X}^{\prime}_{i,j}$, completing the graph generation pipeline.

\subsection{Training Protocol}
\label{sec:training}

The model employs an ``1+2n'' training strategy that builds upon standard pretraining through iterative refinement. The protocol begins with an initial phase focusing on edge reconstruction, where only the edge encoder and decoder components are trained while excluding the Compression and Decompression (CD) module. This stage minimizes the binary edge reconstruction loss $\mathcal{L}_1 = \text{BCEloss}(\mathbf{A}, \mathbf{A}')$ to establish fundamental adjacency matrix reconstruction capability. 

Subsequent phases introduce the CD component while freezing the pretrained edge modules, optimizing two concurrent objectives: the reconstruction loss $\mathcal{L}_2^{\text{rec}} = \text{MSELoss}(\mathbf{X}, \mathbf{X}')$ preserves node feature information during dimensionality reduction, while the KL divergence term $\mathcal{L}_2^{\text{kl}} = D_{\text{KL}}(q_\phi(L|G) \parallel \mathcal{N})$ regularizes the latent space distribution. The complete training cycle then proceeds with full end-to-end optimization using the comprehensive loss function $\mathcal{L}_3 = \mathcal{L}_3^{\text{edge\_rec}} + \mathcal{L}_3^{\text{node\_rec}} + \mathcal{L}_3^{\text{kl}}$, which simultaneously handles edge reconstruction, node classification, and latent space regularization.

The ``1+2n'' training protocol implements an iterative refinement process expressed as $\text{Pretrain}_{\text{edge}} + n \times (\text{Train}_{\text{CD}} + \text{Finetune}_{\text{full}})$.  Through alternating cycles of compression-decompression (CD) training and full model fine-tuning, the framework progressively optimizes the latent representations.  At each iteration, the model adjusts its parameters through coordinated updates to both the encoder and decoder components, systematically improving the balance between reconstruction accuracy and latent space organization.  
The effectiveness of our training methodology will be validated in Appendix~\ref{app:train-mode}.

\subsection{Surrogate model}
To analyze the influence of low-dimensional latent variables $\mathbf{z} \in \mathbb{R}^d$ on graph characteristics, we leverage Shapley Additive exPlanations (SHAP), a game-theoretic approach for model interpretability~\cite{lundberg2017unified}.We provide a brief introduction to the computational principles of SHAP in Appendix~\ref{app:shap}. Since SHAP analysis requires differentiable operations while direct computation of graph statistics is non-differentiable, we employ a surrogate model to approximate the ground-truth graph statistics, enabling the application of SHAP analysis to investigate the relationship between low-dimensional latent representations and the generated graph properties.The surrogate model serves solely as an auxiliary tool for SHAP computation.Moreover, in Section~\ref{sec:shap}, we will validate the strong correlation between fitted values and true values to demonstrate the feasibility of the surrogate model. Notably, we utilize the original ground-truth statistical values rather than the model's approximated outputs for our final analysis.The training methodology for the surrogate models is described as follows:as show in Figure~\ref{fig:1}, for any latent vector $\mathbf{z}$, we first decode it into a generated graph $G = (\mathbf{C}, \mathbf{A})$ using the decoder, then compute both the ground-truth statistics $\mathbf{y} = \big[\text{edge-count}(G), \text{reciprocity}(G), \sum\text{betweenness}(G), \text{non-neuronal}(G)\big]$ and their differentiable approximations $\mathbf{C}' = f_\theta(G)$ through dedicated surrogate models (with distinct architectures for different statistics), optimizing the fidelity via $\mathcal{L}_{\text{surrogate}} = \mathbb{E}\big[\|\mathbf{y} - \mathbf{y}'\|_2^2\big]$.Detailed explanations of the metrics are provided in the Appendix~\ref{app:metrics}.

\paragraph{Edge-count}

To estimate the edge count, we first transform the adjacency matrix $\mathbf{A}$ along the node dimension via $\mathbf{T} = \text{LeakyReLU}(\text{Linear}(\mathbf{A}))$, then flatten and process it through an MLP to obtain the approximated edge count $C'_{\text{edge-count}} = \text{MLP}(\text{flatten}(\mathbf{T}))$. This entire operation is encapsulated in our edge count predictor: $C'_{\text{edge-count}} = f_{\text{edge-count}}(\mathbf{A})$.

\paragraph{Betweenness}
The betweenness centrality summation follows identical architecture to edge count prediction:$C'_{\text{betweenness}} = f_{\text{betweenness}}(\mathbf{A}). $

\paragraph{Reciprocity}

For reciprocity computation, we separately fit the numerator and denominator components using a shared-parameter model architecture $f_\text{reciprocity}$, which maintains the same structure as the edge count predictor. The numerator calculation first applies the transformation $\mathbf{A}_{\text{two}} = \max(\mathbf{A} - 0.5, 0)$ to ensure non-negativity, followed by $C'_{\text{two}} = f_\text{reciprocity}(\mathbf{A}_{\text{two}})$. For the denominator, we compute $\mathbf{A}_{\text{one}} = \max\left((\mathbf{A} + \mathbf{A}^\top - 1)/2, 0\right)$ before obtaining $C'_{\text{one}} = f_\text{reciprocity}(\mathbf{A}_{\text{one}})$. The final reciprocity measure is then derived as:$
C'_{\text{reciprocity}} = \frac{C'_{\text{two}}}{C'_{\text{one}}}.
$

\paragraph{Non-Neuronal}
To compute non-neuronal type counts, we process feature matrix $\mathbf{X}$ as:
\begin{equation}
\mathbf{X}' = \sum_{t \in \mathcal{T}} \max(\mathbf{X}_{\text{NN}} - \mathbf{X}_t, 0), \quad \mathcal{T} = \{\text{GABA}, \text{GLUT}, \text{ACH}, \text{SER}\},
\end{equation}
then obtain $C'_{\text{non-neuronal}} = \text{MLP}(\mathbf{X}')$.

\subsection{Dynamic programming}
The additive property of SHAP values allows us to determine optimal values for each dimension of the low-dimensional representation by analyzing their individual contributions to the target graph properties~\cite{lundberg2017unified}. This enables the generation of graphs whose statistical features approximate our desired specifications. We formulate this optimization task as a knapsack-like problem~\cite{martello1987kanpsack} and solve it efficiently using dynamic programming~\cite{bellman1966dynamic} to obtain the optimal set of latent variables.
As shown in Algorithm~\ref{alg:1}, this approach computes a DP table from which, given any target feature $y$, we can trace back from $\text{dp}[d][y]$ to reconstruct the optimal $\mathbf{Z}$ values that produce graphs with feature $y$.

\begin{algorithm}
\caption{SHAP-Constrained Dynamic Programming}
\label{alg:1}
\begin{algorithmic}[1]
\State \textbf{Input:} 
\State \quad $d$: dimension of $\mathbf{Z} \in \mathbb{R}^d$  
\State \quad $l,r$:possible eigen value bounds ($l \leq \lambda \leq r$)
\State \quad $\text{shap}[i][k]$: SHAP value for $z_i=k$ 
\State \quad $\sigma$: standard deviation of $\mathbf{Z}$
\State \textbf{Output:} DP table $\text{dp}[0..d][l..r]$
\Statex
\State Initialize $\text{dp}[0..d][l..r] \leftarrow -\infty$
\State $\text{dp}[0][0] \leftarrow 0$
\For{$i = 1$ \textbf{to} $d$}
    \For{$j = l$ \textbf{to} $r$}
        \For{$k = -\sigma$ \textbf{to} $\sigma$}
            \If{$\text{dp}[i-1][j] \neq -\infty$}
                \State $\text{dp}[i][j+\text{shap}[i][k]] \leftarrow k$ 
                
            \EndIf
        \EndFor
    \EndFor
\EndFor
\State \Return $\text{dp}$
\end{algorithmic}
\end{algorithm}

\subsection{CMA-ES}
To obtain desired neuronal connectivity patterns through low-dimensional representation adjustment, we employ the Covariance Matrix Adaptation Evolution Strategy (CMA-ES)~\cite{hansen2016cma} to optimize the latent variables $\mathbf{Z}$. This evolutionary algorithm efficiently searches the latent space to find the optimal configuration $\mathbf{Z}^*$ that generates graphs most closely matching our target neural connection structures~\cite{loshchilov2016cmause}. 
The complete technical description of CMA-ES is provided in Appendix~\ref{app:cmaes}.

\section{Experiment}
This study investigates four core research questions(RQ): \textbf{RQ1} evaluates the model's generative capability on graphs,
\textbf{RQ2} examines how low-dimensional latent vectors $\mathbf{z} \in \mathbb{R}^d$ ($d \ll |\mathbf{X}|$) influence target graph features $\mathbf{y}(\mathbf{G}')$, and \textbf{RQ3} develops controlled generation of specified graphs $\mathbf{G}_{\text{target}}$ through optimization-based manipulation of the latent space via $\mathbf{z}^* = \arg\min_{\mathbf{z}} \mathcal{L}(\text{Decoder}(\mathbf{z}), \mathbf{G}_{\text{target}})$.

\subsection{Experimental Setup}
 All experiments can be run on a single NVIDIA-A40 GPU card.
\paragraph{Datasets}
Using the sampling method from Section~\ref{sec:dataset}, we extracted 500 fruit fly neuronal subgraphs $\mathcal{D} = \{G_i\}_{i=1}^{500}$ where each $G_i = (\mathbf{A}_i, \mathbf{X}_i)$ contains an adjacency matrix $\mathbf{A}_i \in \mathbb{R}^{100 \times 100}$ and node features $\mathbf{X}_i \in \mathbb{R}^{100 \times 1}$ representing four neurotransmitter types (GABA, GLUT, ACH, SER) and one non-neuronal type. The dataset was split into training/test/validation sets with an 8:1:1 ratio.
For \textbf{RQ1}, since existing baseline models primarily operate on undirected graphs, we converted the directed connectome graphs to undirected format by symmetrizing the adjacency matrices while preserving all node features. This adaptation ensures fair comparison while maintaining the essential topological properties.

\paragraph{Evaluation}
For \textbf{RQ1}, we employ the same evaluation metrics as Xiaoyang et al.~\cite{hou2024improving} for general graph generation, measuring the Maximum Mean Discrepancy (MMD) between training and generated graphs across three topological features: degree distributions , clustering coefficients , and orbit counts . 

For \textbf{RQ2}, we first validate the \texttt{Surrogate Model} effectiveness by computing Pearson correlation coefficients  between fitted values and ground-truth statistics. We then analyze the relationship between low-dimensional latent vectors and generated graphs through four structural characteristics: edge-count, reciprocity, betweenness, and non-neuronal.

For \textbf{RQ3}, we employ two distinct optimization objectives: (1) direct utilization of the target graph's connectivity information, and (2) matching of the target graph's statistical characteristics, specifically its in-degree and out-degree distributions. The similarity between generated and target graphs is then evaluated using the AUC and accuracy metrics computed between their respective adjacency matrices $\mathbf{A}_{\text{gen}}$ and $\mathbf{A}_{\text{target}}$, where higher values indicate better structural correspondence.
Detailed explanations of the evaluation metrics are provided in the Appendix~\ref{app:metrics}.

\paragraph{Baselines}
For RQ1, we incorporate several state-of-the-art (SOTA) models in graph generation, including \texttt{EDGE}~\cite{chen2023edge}, \texttt{GDSS}~\cite{jo2022GDSS}, \texttt{DisCo}~\cite{xu2024disco}, and \texttt{GruM}~\cite{jo2024GruM}. To ensure maximal fairness in comparison, we randomly sample their hyperparameters within $\pm20\%$ of their original published values. Through 10 independent random sampling trials for each model, we select the best-performing configuration as our baselines.

\subsection{RQ1:Generative capability on FlyWire}

As shown in Table~\ref{table:1}, our method demonstrates competitive performance in generating topological features on the FlyWire dataset. For degree distribution, the model achieves an MMD score of 0.002, matching the current best-performing method GruM and significantly outperforming other baselines. This result confirms that despite the forced low-dimensional compression through the information bottleneck, our model can accurately reconstruct the degree distribution characteristics of real neuronal connectivity.

\begin{table}[!htbp]
\centering
\begin{threeparttable}
\caption{Generation results on FlyWire. We compute the Maximum Mean Discrepancy (MMD) between generated and test graphs for three graph features, with the best results shown in \textbf{bold}.}
\label{table:1}
\begin{tabular}{lcccc}
\toprule
\multirow{2}{*}{\textbf{Method}} & 
\multicolumn{4}{c}{\textbf{FlyWire}}  \\
\cmidrule(lr){2-5} 
& \textbf{Deg.$\downarrow$} & \textbf{Clus.$\downarrow$} & \textbf{Orbit$\downarrow$}  & \textbf{Avg.$\downarrow$}\\
\midrule
EDGE & 0.009 & 0.099 & 0.038 & \textbf{0.048}\\
DisCo & 0.141 & 0.552 & 0.377 & 0.356\\
GDSS & 0.054 & 0.633 & 0.702 & 0.463 \\
GruM & \textbf{0.002} & 0.165 & \textbf{0.035} & 0.067\\
\midrule
Ours & \textbf{0.002} & \textbf{0.056} & 0.129 & 0.062\\
\bottomrule
\end{tabular}
\end{threeparttable}
\end{table}

The model shows particularly strong performance in clustering coefficient, surpassing all comparative methods. This metric reflects the tightness of local circuit connections, and its high fidelity demonstrates our framework's effectiveness in capturing the modular properties of columnar functional units in biological neural networks. In contrast, the reconstruction accuracy for orbit counts is relatively lower, which aligns with expectations: global topological features, being influenced by more stochastic factors, inevitably suffer greater information loss during low-dimensional compression.

\begin{figure}[!htbp]
    \centering
    \begin{subfigure}[b]{0.155\textwidth}
        \includegraphics[width=\linewidth,valign=t]{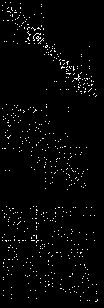}
        \caption{\scriptsize Origin} 
    \end{subfigure}
    \hspace{-0.5em} 
    \begin{subfigure}[b]{0.155\textwidth}
        \includegraphics[width=\linewidth,valign=t]{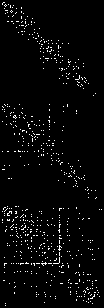}
        \caption{\scriptsize Ours}
    \end{subfigure}
    \hspace{-0.5em}
    \begin{subfigure}[b]{0.155\textwidth}
        \includegraphics[width=\linewidth,valign=t]{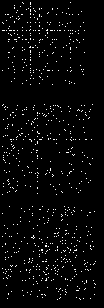}
        \caption{\scriptsize EDGE}
    \end{subfigure}
    \hspace{-0.5em}
    \begin{subfigure}[b]{0.155\textwidth}
        \includegraphics[width=\linewidth,valign=t]{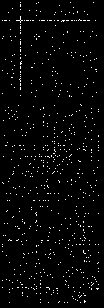}
        \caption{\scriptsize GruM}
    \end{subfigure}
    \hspace{-0.5em}
    \begin{subfigure}[b]{0.155\textwidth}
        \includegraphics[width=\linewidth,valign=t]{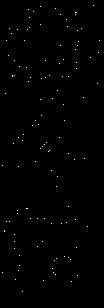}
        \caption{\scriptsize GDSS}
    \end{subfigure}
    \hspace{-0.5em}
    \begin{subfigure}[b]{0.155\textwidth}
        \includegraphics[width=\linewidth,valign=t]{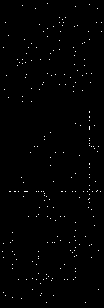}
        \caption{\scriptsize DisCo}
    \end{subfigure}
    
    \caption{Generation results on FlyWire in Adjacency Matrix Format. (a) Original: Shows the sampled graphs from the Drosophila brain connectome with three distinct divergence levels. (b) Ours: Our model's generation results, also exhibiting three divergence levels. (c)-(d) EDGE and GruM results: While performing well on high-divergence graphs, they lack medium and low-divergence samples, indicating limited diversity. (e)-(f) GDSS and DisCo results: Their main limitations are insufficient edge generation and similarly constrained diversity.}
    \label{fig:gen}
\end{figure}

In Figure~\ref{fig:gen}, we present the generation results of our model alongside comparative models on the FlyWire dataset. The results demonstrate that our model exhibits superior performance in generation diversity compared to baseline methods.

This performance differentiation reveals intrinsic characteristics of our method: while the information bottleneck causes slight distortion in global topology, it effectively filters noise and makes key biological patterns (e.g., local clustering) more salient in the latent space. Notably, although EDGE achieves the lowest average MMD, its high-dimensional representation lacks explicit structural constraints, making subsequent causal analysis substantially more challenging to implement. Our method, through careful balancing of reconstruction accuracy and dimensional compression, establishes interpretable mappings between latent variables and generated graph structural features, providing a solid foundation for analyzing brain connectomes.

In addition to this, we also observe the reconstruction performance of our model on FlyWire.For details, please refer to the Appendix~\ref{app:recon}.



\subsection{RQ2:Analysis of low-dimensional variables}
\label{sec:shap}

\begin{table}[h]
\centering
\begin{threeparttable}

\caption{Fitting results of our model on FlyWire. We compute Pearson correlation coefficients between four graph features and their ground-truth statistical values to validate the effectiveness of our surrogate model.}
\label{table:3}
\begin{tabular}{lcccc}
\toprule
\multirow{2}{*}{\textbf{Method}} & 
\multicolumn{4}{c}{\textbf{FlyWire}}  \\
\cmidrule(lr){2-5} 
& \textbf{Edge-Count$\uparrow$} & \textbf{Reciprocity$\uparrow$} & \textbf{Betweenness$\uparrow$}  & \textbf{Non-Neuronal$\uparrow$}\\
\midrule
Ours & 0.9982 & 0.9192 & 0.9622 & 0.9853\\
\bottomrule
\end{tabular}

\end{threeparttable}
\end{table}

\begin{figure}[!htbp]
    \centering
    \begin{subfigure}[t]{0.31\linewidth}  
        \includegraphics[width=\linewidth, height=5cm, keepaspectratio]{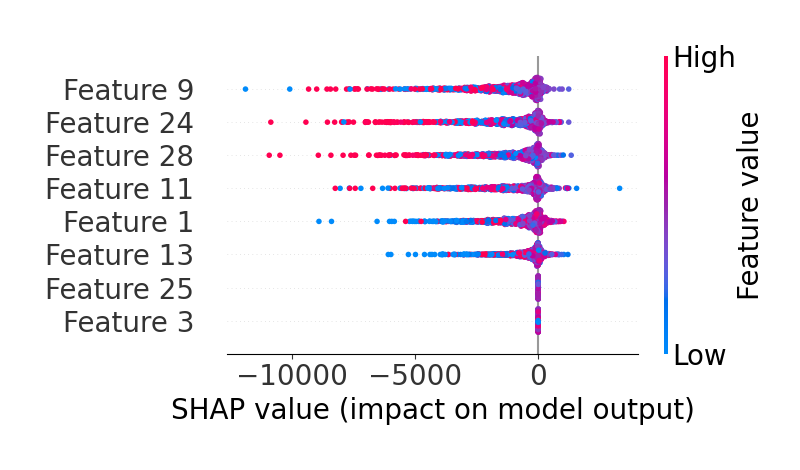}
        \caption{SHAP Values}
        \label{fig:2a}
    \end{subfigure}%
    \hfill
    \begin{subfigure}[t]{0.31\linewidth}
        \includegraphics[width=\linewidth, height=5cm, keepaspectratio]{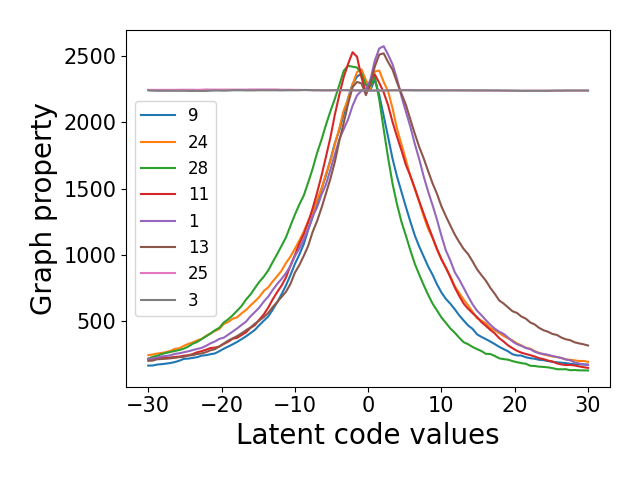}
        \caption{Dimension-Feature Relation}
        \label{fig:2b}
    \end{subfigure}%
    \hfill
    \begin{subfigure}[t]{0.31\linewidth}
        \includegraphics[width=\linewidth, height=5cm, keepaspectratio]{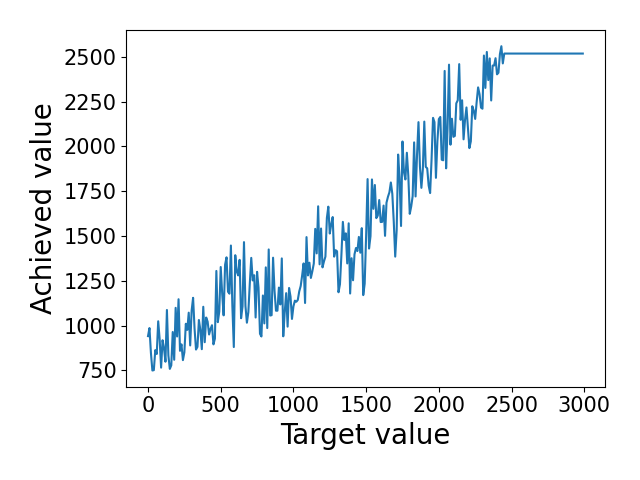}
        \caption{DP Results}
        \label{fig:2c}
    \end{subfigure}
    
    \caption{Analysis of Edge Count.SHAP value results are shown in (a), where the y-axis represents latent dimensions arranged from top (most influential) to bottom (least influential). Point colors indicate magnitude - red for larger values and blue for smaller values. The x-axis shows the direction and strength of impact, with positive values indicating positive influence and negative values indicating negative influence. (b) displays the relationship between each dimension's values and the statistical features of generated graphs. (c) presents the dynamic programming results, with the x-axis showing desired target values and the y-axis showing the actual statistical values of graphs generated from the DP-optimized low-dimensional representations.}
    \label{fig:2}
\end{figure}

We first train our surrogate model to minimize the discrepancy between predicted and ground-truth statistical values. As demonstrated in Table~\ref{table:3}, the Pearson correlation coefficients for all four metrics exceed 0.9, indicating strong agreement between fitted and actual values and validating the surrogate model's effectiveness. Subsequent SHAP analysis reveals the relationship between latent dimensions and graph characteristics (Figure~\ref{fig:2a}). For edge count prediction, only dimensions 1, 9, 11, 13, 24, and 28 exhibit significant influence, with their impact intensifying negatively as values deviate from zero - similar to how a small number of genes can control complex biological structures~\cite{benzer1971fromgene}. In the Appendix~\ref{app:dim}, we provide experimental evidence demonstrating that the remaining dimensions indeed have negligible impact on graph generation. Figure~\ref{fig:2b} illustrates how varying these 8 key dimensions within $\pm3\sigma$ affects the output, confirming our analytical findings. Finally, we employ dynamic programming guided by SHAP values to generate graphs with target features. Figure~\ref{fig:2c} shows a monotonic increase in achieved features versus desired specifications, proving our method's capability for controlled generation.
Other results are provided in the Appendix~\ref{app:other-results}.

\subsection{RQ3:Generating target graphs by adjusting low-dimensional variable}


When utilizing complete information from the target graph, the model achieves an AUC of 0.919 and an accuracy of 97.4\%. Remarkably, even when only employing statistical information of the target graph, it still attains competitive performance with an AUC of 0.852 while maintaining the same accuracy of 97.4\%. These results suggest that specific graph structures can be obtained by adjusting low-dimensional variables. The comparable accuracy rates between these two scenarios may be attributed to the sparsity of edges in the graph topology.

\section{Related Work}

Generative models are a class of artificial intelligence models that learn the distribution of data and generate new samples~\cite{kaswan2023generative}. 
Recent advances in generative models have been dominated by diffusion-based approaches~\cite{zhao2024diffusion,xu2024diffusion,madeira2024diffusion,chen2023diffusion} and flow-based methods~\cite{hou2024flow1,shi2020flow,luo2021flow}. While these demonstrate strong performance in graph generation tasks, they fundamentally lack an information bottleneck mechanism, making them suboptimal for interpretable analysis. 
While existing VAE frameworks can achieve low-dimensional graph representation through the information bottleneck mechanism~\cite{simonovsky2018graphvae}, direct graph-level compression often leads to significant information loss when processing large-scale graph data. To address this limitation, our proposed training method effectively enhances the information retention capacity of graph-level embeddings while preserving the information bottleneck principle.

Disentangled representation learning can acquire semantically meaningful independent representations, which positively contributes to the interpretability of low-dimensional variables. Existing unsupervised VAE-based disentanglement learning methods such as $\beta$-VAE~\cite{higgins2017betavae}, DIP-VAE~\cite{kumar2017dipvae}, and RF-VAE~\cite{kim2019rfvae} promote disentanglement by adjusting regularization terms. However, Michlo et al.~\cite{michlo2022appendence} discovered that the disentanglement achieved through these approaches exhibits coincidental properties, while Locatello et al.~\cite{locatello2019impossible} demonstrated the fundamental difficulty of unsupervised disentanglement. Consequently, we employ SHAP analysis on top of coupled low-dimensional representations~\cite{marcilio2020shapexplanations,ponce2024shappractical}. Although SHAP is conventionally used to interpret input feature importance~\cite{prendin2023shap1,alabi2023shap2,raihan2023shap3,he2025shap4,kim2023shap5}, our approach uniquely combines dynamic programming~\cite{eddy2004dp} with SHAP values to establish a controllable white-box generation process with specified features. Additionally, we utilize CMA-ES to provide a complementary black-box generation approach for targeted graph synthesis~\cite{loshchilov2016cma1,auger2012cma2}.

\section{Conclusion}

This work presents an interpretable framework for connectome analysis using adaptive graph representation learning. Our approach combines functional subgraph sampling with variational autoencoder-based compression to generate compact yet biologically meaningful representations of neuronal networks. The model demonstrates strong performance in both graph reconstruction and feature preservation, as evidenced by high accuracy in node classification and edge prediction tasks. Through SHAP-based analysis, we identify key latent dimensions that govern important structural properties of neural circuits. The proposed "1+2n" training strategy effectively balances reconstruction quality with interpretability while maintaining the information bottleneck principle. Experimental results on the FlyWire dataset validate our method's ability to capture essential organizational patterns in connectomes while remaining computationally efficient. Future work may explore extensions to dynamic connectivity analysis and multi-scale network representations.

\bibliography{main}

\begin{thebibliography}{10}

\bibitem{alabi2023shap2}
Rasheed~Omobolaji Alabi, Mohammed Elmusrati, Ilmo Leivo, Alhadi Almangush, and Antti~A M{\"a}kitie.
\newblock Machine learning explainability in nasopharyngeal cancer survival using lime and shap.
\newblock {\em Scientific Reports}, 13(1):8984, 2023.

\bibitem{auger2012cma2}
Anne Auger and Nikolaus Hansen.
\newblock Tutorial cma-es: evolution strategies and covariance matrix adaptation.
\newblock In {\em Proceedings of the 14th annual conference companion on Genetic and evolutionary computation}, pages 827--848, 2012.

\bibitem{bellman1966dynamic}
Richard Bellman.
\newblock Dynamic programming.
\newblock {\em science}, 153(3731):34--37, 1966.

\bibitem{benzer1971fromgene}
Seymour Benzer.
\newblock From the gene to behavior.
\newblock {\em Jama}, 218(7):1015--1022, 1971.

\bibitem{cai2024neural-architecture2}
Zicheng Cai, Lei Chen, Peng Liu, Tongtao Ling, and Yutao Lai.
\newblock Eg-nas: neural architecture search with fast evolutionary exploration.
\newblock In {\em Proceedings of the AAAI Conference on Artificial Intelligence}, volume~38, pages 11159--11167, 2024.

\bibitem{chen2022data}
Guozhang Chen, Franz Scherr, and Wolfgang Maass.
\newblock A data-based large-scale model for primary visual cortex enables brain-like robust and versatile visual processing.
\newblock {\em Science advances}, 8(44):eabq7592, 2022.

\bibitem{chen2023edge}
Xiaohui Chen, Jiaxing He, Xu~Han, and Li-Ping Liu.
\newblock Efficient and degree-guided graph generation via discrete diffusion modeling.
\newblock {\em arXiv preprint arXiv:2305.04111}, 2023.

\bibitem{chen2023diffusion}
Xiaohui Chen, Jiaxing He, Xu~Han, and Li-Ping Liu.
\newblock Efficient and degree-guided graph generation via discrete diffusion modeling.
\newblock {\em arXiv preprint arXiv:2305.04111}, 2023.

\bibitem{dorkenwald2022flywire}
Sven Dorkenwald, Claire~E McKellar, Thomas Macrina, Nico Kemnitz, Kisuk Lee, Ran Lu, Jingpeng Wu, Sergiy Popovych, Eric Mitchell, Barak Nehoran, et~al.
\newblock Flywire: online community for whole-brain connectomics.
\newblock {\em Nature methods}, 19(1):119--128, 2022.

\bibitem{eddy2004dp}
Sean~R Eddy.
\newblock What is dynamic programming?
\newblock {\em Nature biotechnology}, 22(7):909--910, 2004.

\bibitem{figureau1987genebottle}
A~Figureau.
\newblock Information theory and the genetic code.
\newblock {\em Origins of Life and Evolution of the Biosphere}, 17(3):439--449, 1987.

\bibitem{gao2025snn}
Shouwei Gao, Ruixin Zhu, Yu~Qin, Wenyu Tang, and Hao Zhou.
\newblock Sg-snn: a self-organizing spiking neural network based on temporal information.
\newblock {\em Cognitive Neurodynamics}, 19(1):14, 2025.

\bibitem{guo2022snn2}
Lei Guo, Sijia Zhang, Youxi Wu, and Guizhi Xu.
\newblock Complex spiking neural networks with synaptic time-delay based on anti-interference function.
\newblock {\em Cognitive Neurodynamics}, 16(6):1485--1503, 2022.

\bibitem{hansen2016cma}
Nikolaus Hansen.
\newblock The cma evolution strategy: A tutorial.
\newblock {\em arXiv preprint arXiv:1604.00772}, 2016.

\bibitem{++hassabis2017}
Demis Hassabis, Dharshan Kumaran, Christopher Summerfield, and Matthew Botvinick.
\newblock Neuroscience-inspired artificial intelligence.
\newblock {\em Neuron}, 95(2):245--258, 2017.

\bibitem{he2025shap4}
Xiuling He, Yue Li, Xiong Xiao, Yingting Li, Jing Fang, and Ruijie Zhou.
\newblock Multi-level cognitive state classification of learners using complex brain networks and interpretable machine learning.
\newblock {\em Cognitive Neurodynamics}, 19(1):5, 2025.

\bibitem{higgins2017betavae}
Irina Higgins, Loic Matthey, Arka Pal, Christopher Burgess, Xavier Glorot, Matthew Botvinick, Shakir Mohamed, and Alexander Lerchner.
\newblock beta-vae: Learning basic visual concepts with a constrained variational framework.
\newblock In {\em International conference on learning representations}, 2017.

\bibitem{hou2024improving}
Xiaoyang Hou, Tian Zhu, Milong Ren, Dongbo Bu, Xin Gao, Chunming Zhang, and Shiwei Sun.
\newblock Improving molecular graph generation with flow matching and optimal transport.
\newblock {\em arXiv preprint arXiv:2411.05676}, 2024.

\bibitem{hou2024flow1}
Xiaoyang Hou, Tian Zhu, Milong Ren, Dongbo Bu, Xin Gao, Chunming Zhang, and Shiwei Sun.
\newblock Improving molecular graph generation with flow matching and optimal transport.
\newblock {\em arXiv preprint arXiv:2411.05676}, 2024.

\bibitem{jo2024GruM}
Jaehyeong Jo, Dongki Kim, and Sung~Ju Hwang.
\newblock Graph generation with diffusion mixture.
\newblock {\em arXiv:2302.03596}, 2024.

\bibitem{jo2022GDSS}
Jaehyeong Jo, Seul Lee, and Sung~Ju Hwang.
\newblock Score-based generative modeling of graphs via the system of stochastic differential equations.
\newblock {\em arXiv:2202.02514}, 2022.

\bibitem{kaswan2023generative}
Kuldeep~Singh Kaswan, Jagjit~Singh Dhatterwal, Kiran Malik, and Anupam Baliyan.
\newblock Generative ai: a review on models and applications.
\newblock In {\em 2023 International Conference on Communication, Security and Artificial Intelligence (ICCSAI)}, pages 699--704. IEEE, 2023.

\bibitem{kim2019rfvae}
Minyoung Kim, Yuting Wang, Pritish Sahu, and Vladimir Pavlovic.
\newblock Relevance factor vae: Learning and identifying disentangled factors.
\newblock {\em arXiv preprint arXiv:1902.01568}, 2019.

\bibitem{kim2023shap5}
Seong~Hwan Kim, Hayom Kim, and Jung~Bin Kim.
\newblock Differences in functional network between focal onset nonconvulsive status epilepticus and toxic metabolic encephalopathy: application to machine learning models for differential diagnosis.
\newblock {\em Cognitive Neurodynamics}, 17(4):845--853, 2023.

\bibitem{kingma2013auto}
Diederik~P Kingma, Max Welling, et~al.
\newblock Auto-encoding variational bayes, 2013.

\bibitem{kumar2017dipvae}
Abhishek Kumar, Prasanna Sattigeri, and Avinash Balakrishnan.
\newblock Variational inference of disentangled latent concepts from unlabeled observations.
\newblock {\em arXiv preprint arXiv:1711.00848}, 2017.

\bibitem{lecun2015}
Yann LeCun, Yoshua Bengio, and Geoffrey Hinton.
\newblock Deep learning.
\newblock {\em nature}, 521(7553):436--444, 2015.

\bibitem{liang2024neural-architecture}
Xinyan Liang, Pinhan Fu, Qian Guo, Keyin Zheng, and Yuhua Qian.
\newblock Dc-nas: Divide-and-conquer neural architecture search for multi-modal classification.
\newblock In {\em Proceedings of the AAAI conference on artificial intelligence}, volume~38, pages 13754--13762, 2024.

\bibitem{locatello2019impossible}
Francesco Locatello, Stefan Bauer, Mario Lucic, Gunnar Raetsch, Sylvain Gelly, Bernhard Sch{\"o}lkopf, and Olivier Bachem.
\newblock Challenging common assumptions in the unsupervised learning of disentangled representations.
\newblock In {\em international conference on machine learning}, pages 4114--4124. PMLR, 2019.

\bibitem{loshchilov2016cmause}
Ilya Loshchilov and Frank Hutter.
\newblock Cma-es for hyperparameter optimization of deep neural networks.
\newblock {\em arXiv preprint arXiv:1604.07269}, 2016.

\bibitem{loshchilov2016cma1}
Ilya Loshchilov and Frank Hutter.
\newblock Cma-es for hyperparameter optimization of deep neural networks.
\newblock {\em arXiv preprint arXiv:1604.07269}, 2016.

\bibitem{lundberg2017unified}
Scott~M Lundberg and Su-In Lee.
\newblock A unified approach to interpreting model predictions.
\newblock {\em Advances in neural information processing systems}, 30, 2017.

\bibitem{luo2021flow}
Youzhi Luo, Keqiang Yan, and Shuiwang Ji.
\newblock Graphdf: A discrete flow model for molecular graph generation.
\newblock In {\em International conference on machine learning}, pages 7192--7203. PMLR, 2021.

\bibitem{madeira2024diffusion}
Manuel Madeira, Clement Vignac, Dorina Thanou, and Pascal Frossard.
\newblock Generative modelling of structurally constrained graphs.
\newblock {\em Advances in Neural Information Processing Systems}, 37:137218--137262, 2024.

\bibitem{marblestone2016}
Adam~H Marblestone, Greg Wayne, and Konrad~P Kording.
\newblock Toward an integration of deep learning and neuroscience.
\newblock {\em Frontiers in computational neuroscience}, 10:94, 2016.

\bibitem{marcilio2020shapexplanations}
Wilson~E Marc{\'\i}lio and Danilo~M Eler.
\newblock From explanations to feature selection: assessing shap values as feature selection mechanism.
\newblock In {\em 2020 33rd SIBGRAPI conference on Graphics, Patterns and Images (SIBGRAPI)}, pages 340--347. Ieee, 2020.

\bibitem{martello1987kanpsack}
Silvano Martello and Paolo Toth.
\newblock Algorithms for knapsack problems.
\newblock {\em North-Holland Mathematics Studies}, 132:213--257, 1987.

\bibitem{michlo2022appendence}
Nathan Michlo, Richard Klein, and Steven James.
\newblock Overlooked implications of the reconstruction loss for vae disentanglement.
\newblock {\em arXiv preprint arXiv:2202.13341}, 2022.

\bibitem{ostojic2024computational}
Srdjan Ostojic and Stefano Fusi.
\newblock Computational role of structure in neural activity and connectivity.
\newblock {\em Trends in Cognitive Sciences}, 2024.

\bibitem{ponce2024shappractical}
Ana~Victoria Ponce-Bobadilla, Vanessa Schmitt, Corinna~S Maier, Sven Mensing, and Sven Stodtmann.
\newblock Practical guide to shap analysis: explaining supervised machine learning model predictions in drug development.
\newblock {\em Clinical and Translational Science}, 17(11):e70056, 2024.

\bibitem{prendin2023shap1}
Francesco Prendin, Jacopo Pavan, Giacomo Cappon, Simone Del~Favero, Giovanni Sparacino, and Andrea Facchinetti.
\newblock The importance of interpreting machine learning models for blood glucose prediction in diabetes: an analysis using shap.
\newblock {\em Scientific reports}, 13(1):16865, 2023.

\bibitem{raihan2023shap3}
Md~Johir Raihan, Md~Al-Masrur Khan, Seong-Hoon Kee, and Abdullah-Al Nahid.
\newblock Detection of the chronic kidney disease using xgboost classifier and explaining the influence of the attributes on the model using shap.
\newblock {\em Scientific Reports}, 13(1):6263, 2023.

\bibitem{seung2012}
Sebastian Seung.
\newblock {\em Connectome: How the brain's wiring makes us who we are}.
\newblock HMH, 2012.

\bibitem{shi2020flow}
Chence Shi, Minkai Xu, Zhaocheng Zhu, Weinan Zhang, Ming Zhang, and Jian Tang.
\newblock Graphaf: a flow-based autoregressive model for molecular graph generation.
\newblock {\em arXiv preprint arXiv:2001.09382}, 2020.

\bibitem{shuvaev2024pnas}
Sergey Shuvaev, Divyansha Lachi, Alexei Koulakov, and Anthony Zador.
\newblock Encoding innate ability through a genomic bottleneck.
\newblock {\em Proceedings of the National Academy of Sciences}, 121(38):e2409160121, 2024.

\bibitem{simonovsky2018graphvae}
Martin Simonovsky and Nikos Komodakis.
\newblock Graphvae: Towards generation of small graphs using variational autoencoders.
\newblock In {\em Artificial Neural Networks and Machine Learning--ICANN 2018: 27th International Conference on Artificial Neural Networks, Rhodes, Greece, October 4-7, 2018, Proceedings, Part I 27}, pages 412--422. Springer, 2018.

\bibitem{sporns2011}
Olaf Sporns, MARK DALEY, and JODY~C CULHAM.
\newblock Networks of the brain.
\newblock {\em Canadian Psychology}, 52(4):321--322, 2011.

\bibitem{sun2025snn1}
Hongze Sun, Shifeng Mao, Wuque Cai, Yan Cui, Duo Chen, Dezhong Yao, and Daqing Guo.
\newblock Bisnn: bio-information-fused spiking neural networks for enhanced eeg-based emotion recognition.
\newblock {\em Cognitive Neurodynamics}, 19(1):1--12, 2025.

\bibitem{velivckovic2017graph}
Petar Veli{\v{c}}kovi{\'c}, Guillem Cucurull, Arantxa Casanova, Adriana Romero, Pietro Lio, and Yoshua Bengio.
\newblock Graph attention networks.
\newblock {\em arXiv preprint arXiv:1710.10903}, 2017.

\bibitem{xu2024disco}
Zhe Xu, Ruizhong Qiu, Yuzhong Chen, Huiyuan Chen, Xiran Fan, Menghai Pan, Zhichen Zeng, Mahashweta Das, and Hanghang Tong.
\newblock Discrete-state continuous-time diffusion for graph generation.
\newblock {\em arXiv preprint arXiv:2405.11416}, 2024.

\bibitem{xu2024diffusion}
Zhe Xu, Ruizhong Qiu, Yuzhong Chen, Huiyuan Chen, Xiran Fan, Menghai Pan, Zhichen Zeng, Mahashweta Das, and Hanghang Tong.
\newblock Discrete-state continuous-time diffusion for graph generation.
\newblock {\em arXiv preprint arXiv:2405.11416}, 2024.

\bibitem{zador2019}
Anthony~M Zador.
\newblock A critique of pure learning and what artificial neural networks can learn from animal brains.
\newblock {\em Nature communications}, 10(1):3770, 2019.

\bibitem{zhao2024diffusion}
Lingxiao Zhao, Xueying Ding, and Leman Akoglu.
\newblock Pard: Permutation-invariant autoregressive diffusion for graph generation.
\newblock {\em arXiv preprint arXiv:2402.03687}, 2024.

\end{thebibliography}


\appendix

\section{Graph Attention Network}
\label{app:gat}

The Graph Attention Network (GAT) employs an attention mechanism to dynamically capture structural relationships in graph data. Unlike conventional graph convolutional networks, GAT computes adaptive attention weights between connected nodes, allowing each node to selectively aggregate features from its most relevant neighbors. Given node features $\vec{h}_i$ and a shared weight matrix $\mathbf{W}$, the attention mechanism operates through three sequential computations:

First, the unnormalized attention coefficient between nodes $i$ and $j$ is calculated as:
\begin{equation}
e_{ij} = \text{LeakyReLU}\left(\mathbf{a}^\top [\mathbf{W}\vec{h}_i \| \mathbf{W}\vec{h}_j]\right),
\end{equation}
where $\mathbf{a}$ is a learnable attention vector and $\|$ denotes concatenation. These coefficients are then normalized across each node's neighborhood $\mathcal{N}_i$ using softmax:
\begin{equation}
\alpha_{ij} = \frac{\exp(e_{ij})}{\sum_{k \in \mathcal{N}_i} \exp(e_{ik})}.
\end{equation}
The final node representation $\vec{h}_i'$ is obtained by weighted aggregation with nonlinear activation $\sigma$:
\begin{equation}
\vec{h}_i' = \sigma\left(\sum_{j \in \mathcal{N}_i} \alpha_{ij} \mathbf{W} \vec{h}_j\right).
\end{equation}

This architecture provides two key advantages: (1) implicit handling of heterogeneous node degrees through adaptive neighborhood weighting, and (2) automatic learning of edge importance without requiring pre-defined edge weights. The attention mechanism is typically extended to multi-head configurations for enhanced representational capacity.

\section{VAE}
\label{app:vae}
The Variational Autoencoder (VAE) consists of an encoder $\phi$ and decoder $\theta$. Its objective is to maximize the log-likelihood function $\log p(x;\theta)$ of the generated graphs. By introducing the variational posterior distribution $q_\phi(z|x)$, we can transform $\log p(x;\theta)$ as follows:
\begin{equation}
\begin{aligned}
\log p(x;\theta) & =\int_zq(z|x)\log p(x)dz \\
 & =\int_zq(z|x;\phi)\log(\frac{p(z,x)}{p(z|x)})dz \\
 & =\int_zq(z|x;\phi)\log(\frac{p(z,x)}{q(z|x;\phi)}\frac{q(z|x;\phi)}{p(z|x)})dz \\
 & =\int_zq(z|x;\phi)\log(\frac{p(z,x)}{q(z|x;\phi)})dz+\int_zq(z|x;\phi)\log(\frac{q(z|x;\phi)}{p(z|x)})dz \\
 & =\int_zq(z|x;\phi)\log(\frac{p(x|z;\theta)p(z)}{q(z|x;\phi)})dz+\int_zq(z|x;\phi)\log(\frac{q(z|x;\phi)}{p(z|x)})dz \\
 & =ELBO(q, x ; \theta, \phi)+KL(q(z|x;\phi)||p(z|x)).
\end{aligned}
\end{equation}
By rearranging the above equation, we obtain:
\begin{equation}
    \mathrm{KL}\bigl(q(\mathbf{z}\mid\mathbf{x};\boldsymbol{\phi})\bigm\Vert p(\mathbf{z}\mid\mathbf{x})\bigr) = \log p(\mathbf{x};\boldsymbol{\theta}) - \mathrm{ELBO}.
\end{equation}

It can be seen that when the encoder remains unchanged, to maximize $\log p(\mathbf{x};\boldsymbol{\theta})$, we only need to maximize the ELBO. Decomposing the ELBO yields:
\begin{equation}
\begin{aligned}
\max _{\theta, \phi} E L B O(q, x ; \theta, \phi) & =\max _{\theta, \phi} \int_{z} q(z \mid x ; \phi) \log \left(p(x \mid z ; \theta) \frac{p(z)}{q(z \mid x ; \phi)}\right) d z \\
& =\max _{\theta, \phi} \int_{z} q(z \mid x ; \phi) \log (p(x \mid z ; \theta)) d z\\
&+\int_{z} q(z \mid x ; \phi) \log \left(\frac{p(z)}{q(z \mid x ; \phi)}\right) d z \\
& =\max _{\theta, \phi} \mathbb{E}_{z \sim q(z \mid x ; \phi)}[\log p(x \mid z ; \theta)]-K L(q(z \mid x ; \phi), p(z)),
\end{aligned}
\end{equation}
where the first term corresponds to minimizing the reconstruction error, and the second term acts as a regularization term.

\section{Covariance Matrix Adaptation Evolution Strategy (CMA-ES)}
\label{app:cmaes}
The Covariance Matrix Adaptation Evolution Strategy (CMA-ES) is a  evolutionary algorithm for black-box optimization in continuous domains, characterized by its adaptive adjustment of the covariance matrix of a multivariate normal search distribution. At generation $t$, the algorithm samples $\lambda$ candidate solutions $\mathbf{x}_i \sim \mathcal{N}(\mathbf{m}_t, \sigma_t^2\mathbf{C}_t)$ where $\mathbf{m}_t$ represents the mean vector, $\mathbf{C}_t$ the covariance matrix, and $\sigma_t$ the step-size. The mean vector undergoes weighted recombination: $\mathbf{m}_{t+1} = \sum_{i=1}^{\mu} w_i \mathbf{x}_{i:\lambda}$, with $\mu$ denoting the number of elite solutions, $w_i$ their weight coefficients, and $\mathbf{x}_{i:\lambda}$ the $i$-th best solution ranked by fitness. The covariance matrix update ingeniously combines rank-$\mu$ and rank-1 updates through $\mathbf{C}_{t+1} = (1-c_1-c_\mu)\mathbf{C}_t + c_1\mathbf{p}_c\mathbf{p}_c^\top + c_\mu \sum_{i=1}^\mu w_i \mathbf{y}_{i:\lambda}\mathbf{y}_{i:\lambda}^\top$, where $\mathbf{y}_{i:\lambda} = (\mathbf{x}_{i:\lambda}-\mathbf{m}_t)/\sigma_t$ constitutes the standardized mutation vector and $\mathbf{p}_c$ the evolution path accumulator. Concurrently, the step-size $\sigma_t$ adapts via the conjugate evolution path $\mathbf{p}_\sigma$ using $\sigma_{t+1} = \sigma_t \exp\left(\frac{c_\sigma}{d_\sigma}\left(\frac{\|\mathbf{p}_\sigma\|}{\mathbb{E}\|\mathcal{N}(0,\mathbf{I})\|} - 1\right)\right)$. This unique covariance matrix learning mechanism enables CMA-ES to automatically detect dependencies between variables in the objective function while dynamically balancing exploration and exploitation during the search process, making it particularly effective for ill-conditioned and non-separable problems.

\section{Metrics}
\label{app:metrics}
\paragraph{Maximum Mean Discrepancy (MMD)}

The Maximum Mean Discrepancy (MMD) measures the distance between probability distributions by embedding them in a reproducing kernel Hilbert space (RKHS) $\mathcal{H}_k$. For distributions $P$ and $Q$ over $\mathcal{X}$ and a characteristic kernel $k: \mathcal{X} \times \mathcal{X} \rightarrow \mathbb{R}$, the MMD is defined through the following supremum:

\begin{equation}
\mathrm{MMD}(P,Q) = \sup_{\substack{f \in \mathcal{H}_k \\ \|f\|_{\mathcal{H}_k} \leq 1}} \left| \mathbb{E}_{x \sim P}[f(x)] - \mathbb{E}_{y \sim Q}[f(y)] \right|.
\end{equation}

This supremum admits a closed-form expression via kernel mean embeddings:

\begin{equation}
\mathrm{MMD}^2(P,Q) = \|\mu_P - \mu_Q\|_{\mathcal{H}_k}^2 = \mathbb{E}_{x,x' \sim P}[k(x,x')] + \mathbb{E}_{y,y' \sim Q}[k(y,y')] - 2\mathbb{E}_{x \sim P, y \sim Q}[k(x,y)],
\end{equation}

where $\mu_P = \mathbb{E}_{x \sim P}[\phi(x)]$ and $\mu_Q = \mathbb{E}_{y \sim Q}[\phi(y)]$ are the kernel mean embeddings, with $\phi(x) = k(x,\cdot)$ being the canonical feature map. Given two sets of samples $X = \{x_i\}_{i=1}^m \sim P$ and $Y = \{y_j\}_{j=1}^n \sim Q$, the unbiased empirical estimate is computed as:

\begin{equation}
\widehat{\mathrm{MMD}}^2(X,Y) = \frac{1}{m(m-1)} \sum_{i \neq j} k(x_i,x_j) + \frac{1}{n(n-1)} \sum_{i \neq j} k(y_i,y_j) - \frac{2}{mn} \sum_{i,j} k(x_i,y_j).
\end{equation}

The choice of kernel $k$ determines the properties of the metric. The Gaussian kernel $k_\sigma(x,y) = \exp\left(-\frac{\|x-y\|^2}{2\sigma^2}\right)$ is commonly used due to its universality. 

\paragraph{Degree distribution}
The degree distribution $P(k)$ of a graph $G=(V,E)$ formally characterizes the probability distribution of vertex degrees over the entire graph structure. For a graph with $n=|V|$ vertices, the degree $k_i$ of vertex $v_i \in V$ counts its adjacent edges:

\begin{equation}
k_i = \sum_{j=1}^n A_{ij},
\end{equation}

where $A$ is the adjacency matrix satisfying $A_{ij} = 1$ if $(v_i,v_j) \in E$ and $0$ otherwise.

\paragraph{Clustering coefficient}

The clustering coefficient quantifies the tendency of nodes to form tightly connected neighborhoods in a graph $G=(V,E)$. For an undirected graph with adjacency matrix $A$, we compute the clustering coefficient as:

For a vertex $v_i \in V$ with degree $k_i \geq 2$, its local clustering coefficient $C_i$ measures the fraction of possible triangles that exist:

\begin{equation}
C_i = \frac{2T_i}{k_i(k_i-1)} = \frac{\sum_{j,m} A_{ij}A_{jm}A_{mi}}{k_i(k_i-1)},
\end{equation}

where $T_i$ counts the number of triangles incident to vertex $v_i$, and $k_i(k_i-1)/2$ represents the maximum possible triangles. For $k_i \in \{0,1\}$, we define $C_i = 0$.

\paragraph{Orbit counts}

Orbit counts provide a refined topological characterization of graphs by enumerating node positions within small induced subgraphs. For a graph \( G = (V, E) \), the orbit count \( o_i(k) \) of node \( v_i \in V \) in orbit \( k \) is computed as:

\begin{equation}
o_i(k) = \sum_{S \subseteq V, |S| = s} \mathbb{I}\big[v_i \in S \text{ and } v_i \text{ occupies orbit } k \text{ in } G[S]\big],
\end{equation}

where \( G[S] \) denotes the subgraph induced by node set \( S \) of size \( s \) (typically \( s = 3, 4 \)), and \( \mathbb{I}[\cdot] \) is the indicator function(The value is 1 when the condition is satisfied, otherwise 0). 

\paragraph{Edge-count}
Let \( G = (V, E) \) be a graph with vertex set \( V \) and edge set \( E \). The total edge count \( \tau(G) \) is the cardinality of the edge set:

\begin{equation}
\tau(G) = |E| .
\end{equation}
\paragraph{Reciprocity}

Let $G=(V,E)$ be a directed graph with adjacency matrix $A\in\{0,1\}^{|V|\times|V|}$. The reciprocity index $\rho(G)$ quantifying bidirectional connectivity is defined as:

\begin{equation}
\rho(G) = \frac{\sum_{i<j} A_{ij}A_{ji}}{\sum_{i\neq j} A_{ij}(1-A_{ji})} = \frac{\mathrm{tr}(A^2)}{\|A\|_0 - \mathrm{tr}(A^2)},
\end{equation}

where $\mathrm{tr}(A^2)$ counts bidirectional edges and $\|A\|_0$ is the total edge count. For neural networks, this measures feedback complexity through the ratio of reciprocal connections ($A_{ij}=A_{ji}=1$) to unidirectional connections ($A_{ij}=1 \land A_{ji}=0$). 

For convenience in analysis, we scale the actual computation by a factor of 1000.
\paragraph{Betweenness}

The betweenness centrality $c_B(v)$ of a node $v \in V$ quantifies its influence over information flow by measuring the fraction of shortest paths passing through it. Formally, for a graph $G=(V,E)$, the metric is defined as:

\begin{equation}
c_B(v) = \sum_{s \neq v \neq t \in V} \frac{\sigma_{st}(v)}{\sigma_{st}},
\end{equation}

where $\sigma_{st}$ denotes the total number of shortest paths between nodes $s$ and $t$, and $\sigma_{st}(v)$ counts those paths traversing $v$. For neural networks, this captures information routing efficiency through the synaptic topology. Since the normalized centrality values lie within $[0,1]$, we compute the total betweenness centrality as the aggregate measure across all nodes:

\begin{equation}
C_B(G) = \sum_{v \in V} c'_B(v).
\end{equation}

This summation characterizes the global capacity for efficient information routing through the neural architecture, with higher values indicating greater potential for parallel signal propagation across multiple central nodes.

In practical computations, we scale the value by a factor of 100 to mitigate error propagation.

\paragraph{Non-Neuronal}
In our fixed-size graph, let $G = (V, E)$ be a padded graph with $n = |V|$ vertices, where a subset $V_{\text{real}} \subset V$ constitutes the original nodes and $V_{\text{pad}} = V \setminus V_{\text{real}}$ denotes the padding nodes introduced to achieve uniform dimensionality. The non-neuronal type count $\eta(G)$ is defined as the cardinality of the padding set:

\begin{equation}
\eta(G) = |V_{\text{pad}}| = n - |V_{\text{real}}|.
\end{equation}

This metric quantifies the degree of artificial modification needed to conform to the architectural constraints, with higher values suggesting greater distortion from the original topology. The proportion of valid nodes is consequently given by $1 - \frac{\eta(G)}{n}$, which reflects the graph's representation fidelity under padding.
In practical computations, we scale the value by a factor of 100 to mitigate error propagation.

\section{Shap}
\label{app:shap}
The SHAP framework provides a unified approach for interpreting machine learning model predictions through coalitional game theory. Given a trained model $f: \mathbb{R}^d \rightarrow \mathbb{R}$ and an input instance $\mathbf{x} \in \mathbb{R}^d$, the SHAP value $\phi_j$ for feature $j$ represents its marginal contribution to the prediction $f(\mathbf{x})$, computed as:

\begin{equation}
\phi_j(f, \mathbf{x}) = \sum_{S \subseteq \{1,\ldots,d\} \setminus \{j\}} \frac{|S|!(d - |S| - 1)!}{d!} \left( f_{S \cup \{j\}}(\mathbf{x}_{S \cup \{j\}}) - f_S(\mathbf{x}_S) \right),
\end{equation}

where $S$ denotes feature subsets, $|S|$ is the subset cardinality, and $f_S$ represents the conditional expectation of the model output when only features in $S$ are observed:

\begin{equation}
f_S(\mathbf{x}_S) = \mathbb{E}[f(\mathbf{x}) | \mathbf{x}_S] = \int f(\mathbf{x}_S, \mathbf{x}_{\overline{S}}) p(\mathbf{x}_{\overline{S}} | \mathbf{x}_S) d\mathbf{x}_{\overline{S}}.
\end{equation}

The SHAP values satisfy the efficiency property ensuring exact additive decomposition of predictions:

\begin{equation}
f(\mathbf{x}) - \mathbb{E}[f] = \sum_{j=1}^d \phi_j(f, \mathbf{x}).
\end{equation}

For neural networks, the conditional expectations are approximated using background samples $\{\mathbf{z}_k\}_{k=1}^K$ drawn from the training distribution:

\begin{equation}
f_S(\mathbf{x}_S) \approx \frac{1}{K} \sum_{k=1}^K f(\mathbf{x}_S, \mathbf{z}_{k,\overline{S}}).
\end{equation}

The magnitude $|\phi_j|$ indicates feature importance, while the sign distinguishes positive/negative contributions to the prediction. This Shapley-value-based interpretation maintains local accuracy (matching the original model output) and consistency (feature attribution changes monotonically with model dependence).

\section{Effectiveness of 1+2n Training Strategy}
\label{app:train-mode}

\begin{figure}[h]
	\begin{minipage}{0.49\linewidth}
		\vspace{2pt}
		\centerline{\includegraphics[width=\textwidth]{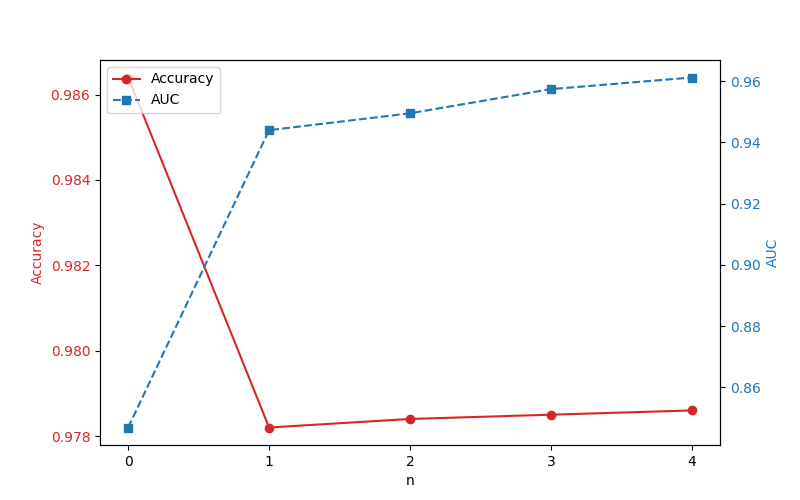}}
		\centerline{(a) The Results of Edge}
	\end{minipage}
	\begin{minipage}{0.49\linewidth}
		\vspace{2pt}
		\centerline{\includegraphics[width=\textwidth]{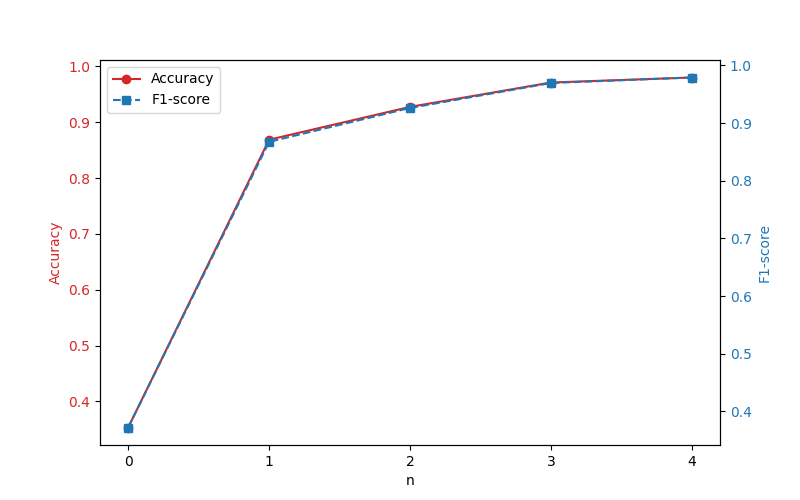}}
		\centerline{(b) The Results of Node}
	\end{minipage}
 
	\caption{Reconstruction Capability Versus N.Figure (a) demonstrates the variation of edge reconstruction metrics with changing values of n, while figure (b) shows the corresponding changes in node category recovery performance metrics as n changes.}
	\label{fig:3}
\end{figure}

In this section, we investigate the efficacy of the proposed 1+2n training methodology. We conduct a series of experiments evaluating the model's performance on directed graphs with $n=0,1,2,3,4$, where $n=0$ represents conventional training without the 1+2n strategy. We primarily focus on evaluating the model's reconstruction capability, as it is more straightforward to assess and more intuitive to interpret compared to generation performance. Specifically, we evaluate edge reconstruction performance using both the Area Under the Curve (AUC) and classification accuracy metrics to compare the original and reconstructed adjacency matrices, while node category recovery is assessed through F1-score and classification accuracy measurements between the original and predicted node categories. 

As demonstrated in Figure~\ref{fig:3}, \texttt{edge auc}, \texttt{node acc} and \texttt{node f1-score} exhibit monotonically increasing trends with larger $n$ values, while \texttt{node acc} shows consistent improvement. Notably, even the minimal case ($n=1$) achieves significant performance gains compared to the baseline ($n=0$). However, for \texttt{edge acc}, although it gradually increases starting from $n=1$, the baseline ($n=0$) surprisingly outperforms other configurations by a slight margin. This phenomenon may stem from the inherent edge sparsity in the graph data. 

The empirical evidence confirms that our 1+2n training strategy effectively enhances graph reconstruction capability.

\section{Dimensionality Analysis of Latent Variables}
\label{app:dim}

\begin{figure}[h]
	\begin{minipage}{0.49\linewidth}
		\vspace{2pt}
		\centerline{\includegraphics[width=\textwidth]{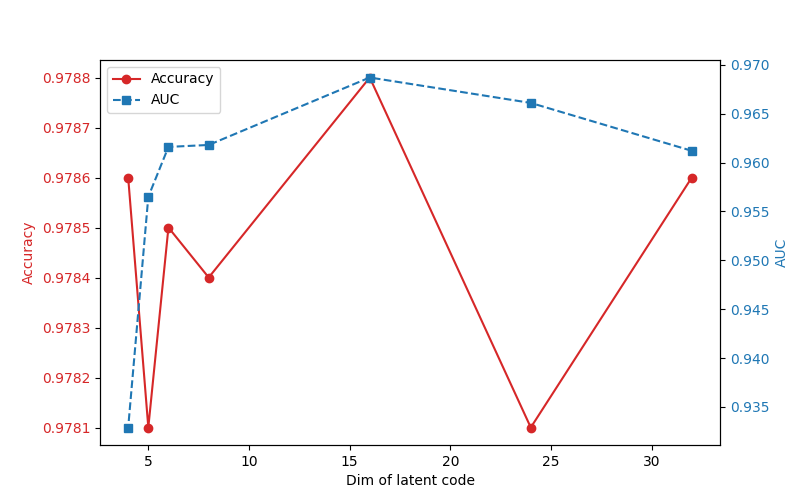}}
		\centerline{(a) The Results of Edge}
	\end{minipage}
	\begin{minipage}{0.49\linewidth}
		\vspace{2pt}
		\centerline{\includegraphics[width=\textwidth]{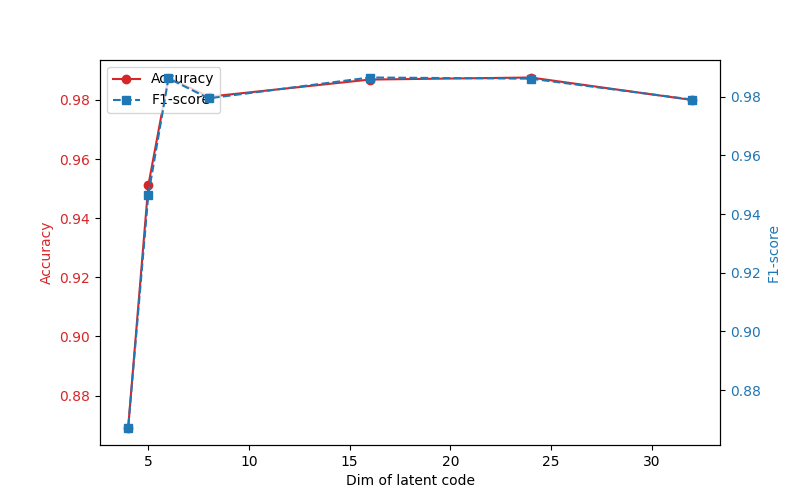}}
		\centerline{(b) The Results of Node}
	\end{minipage}
 
	\caption{Model Reconstruction Capacity under Latent Code Variation.Figure (a) illustrates the variation of edge reconstruction metrics with respect to changes in latent code dimensionality, while figure (b) displays the corresponding changes in node category restoration metrics across different latent code dimensionality.}
	\label{fig:4}
\end{figure}

This section investigates the impact of latent space dimensionality on graph reconstruction capability.We primarily focus on evaluating the model's reconstruction capability, as it is more straightforward to assess and more intuitive to interpret compared to generation performance. As shown in Figure~\ref{fig:4}, the reconstruction capability for both edges and node categories remains stable when reducing the dimensionality from 32 to 6. However, at dimension 5, we observe a slight degradation in edge prediction AUC  accompanied by significant drops in both node classification f1-score  and accuracy. Further reduction to 4 dimensions exacerbates these effects. Notably, edge prediction accuracy remains unaffected across all dimensionalities, likely due to the inherent sparsity of connections in the graph topology.

\section{Additional experimental results}
\label{app:other-results}

\begin{figure}[h]
	
	\begin{minipage}{0.32\linewidth}
		\vspace{3pt}
		\centerline{\includegraphics[width=\textwidth]{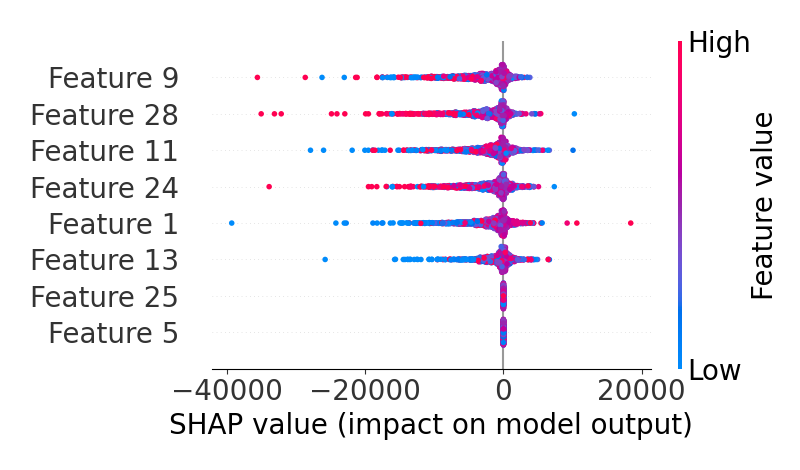}}
		\centerline{(a) Betweenness}
	\end{minipage}
	\begin{minipage}{0.32\linewidth}
		\vspace{3pt}
		\centerline{\includegraphics[width=\textwidth]{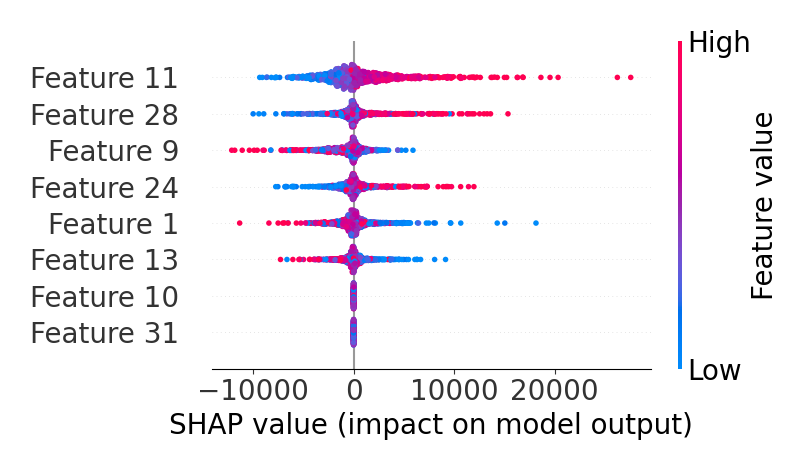}}
	 
		\centerline{(b) Non-Neuronal}
	\end{minipage}
	\begin{minipage}{0.32\linewidth}
		\vspace{3pt}
		\centerline{\includegraphics[width=\textwidth]{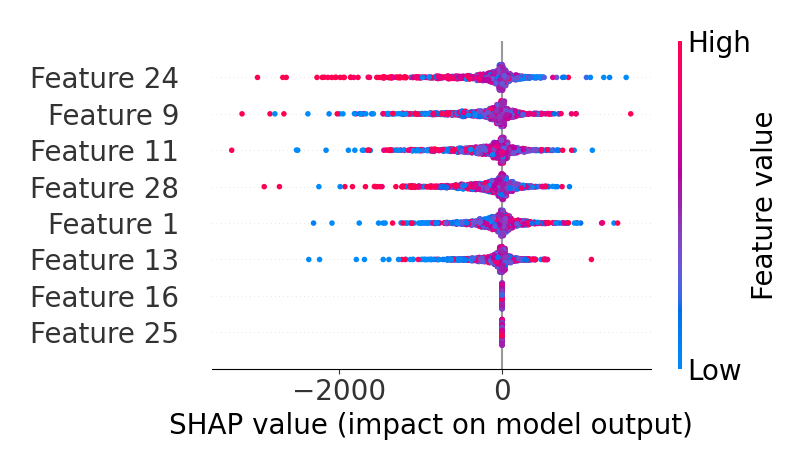}}
	 
		\centerline{(c) Reciprocity}
	\end{minipage}
 
	\caption{SHAP Values.}
	\label{fig:5}
\end{figure}

\begin{figure}[h]
	
	\begin{minipage}{0.32\linewidth}
		\vspace{3pt}
		\centerline{\includegraphics[width=\textwidth]{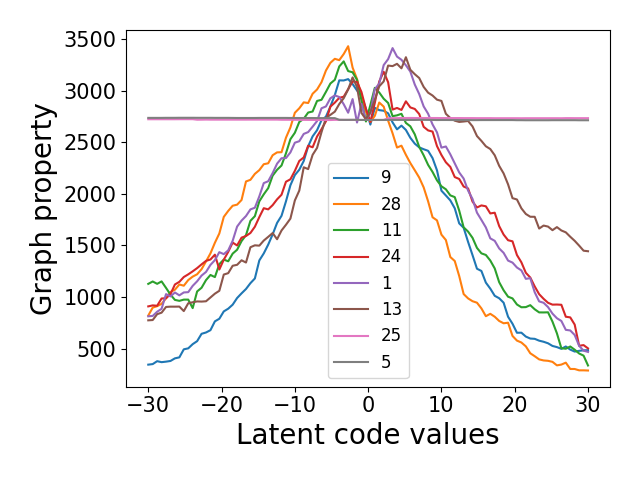}}
		\centerline{(a) Betweenness}
	\end{minipage}
	\begin{minipage}{0.32\linewidth}
		\vspace{3pt}
		\centerline{\includegraphics[width=\textwidth]{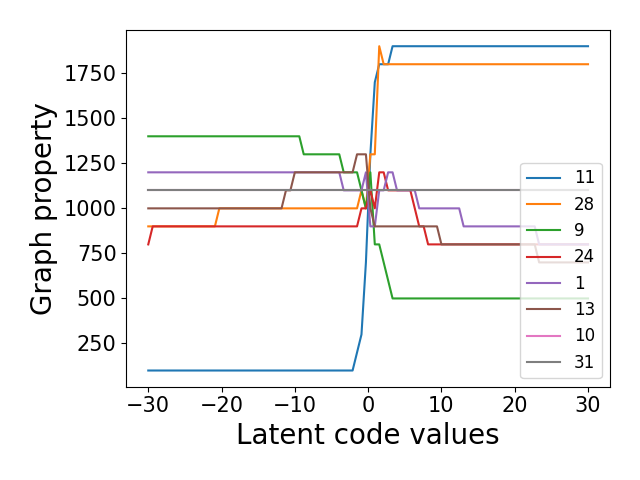}}
	 
		\centerline{(b) Non-Neuronal}
	\end{minipage}
	\begin{minipage}{0.32\linewidth}
		\vspace{3pt}
		\centerline{\includegraphics[width=\textwidth]{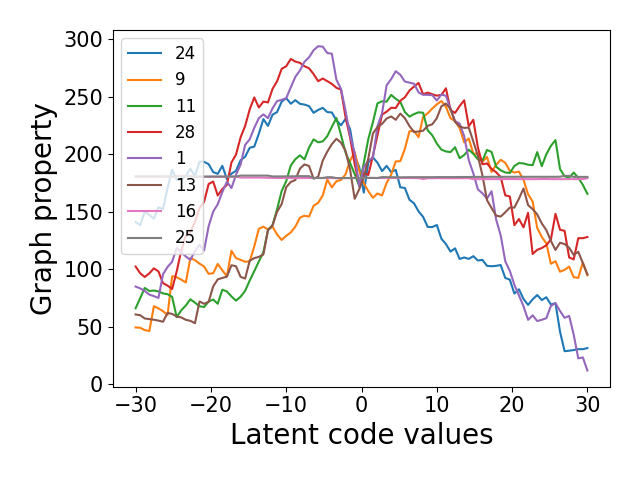}}
	 
		\centerline{(c) Reciprocity}
	\end{minipage}
 
	\caption{Dimension-Feature Relationships.}
	\label{fig:6}
\end{figure}

\begin{figure}[!htbp]
	
	\begin{minipage}{0.32\linewidth}
		\vspace{3pt}
		\centerline{\includegraphics[width=\textwidth]{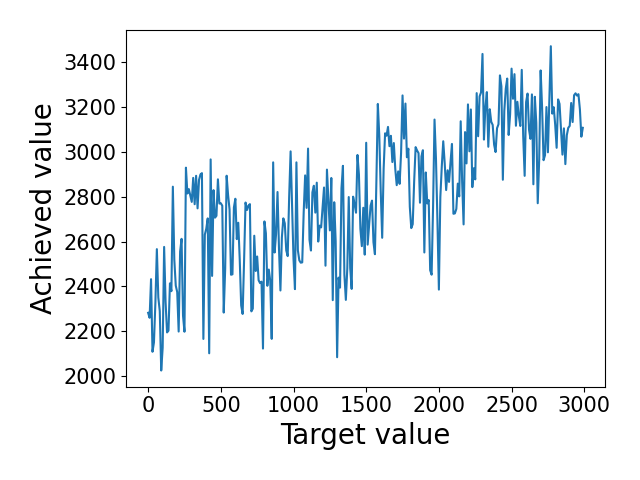}}
		\centerline{(a) Betweenness}
	\end{minipage}
	\begin{minipage}{0.32\linewidth}
		\vspace{3pt}
		\centerline{\includegraphics[width=\textwidth]{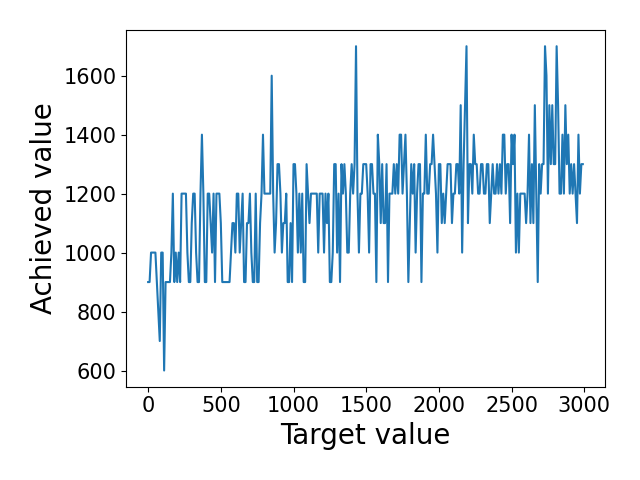}}
	 
		\centerline{(b) Non-Neuronal}
	\end{minipage}
	\begin{minipage}{0.32\linewidth}
		\vspace{3pt}
		\centerline{\includegraphics[width=\textwidth]{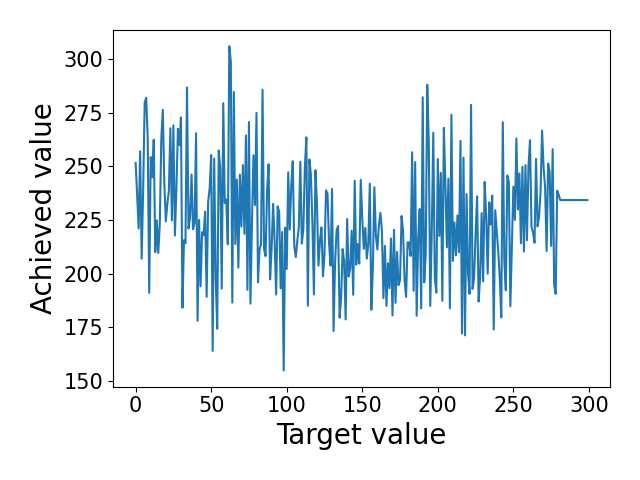}}
	 
		\centerline{(c) Reciprocity}
	\end{minipage}
 
	\caption{DP Results.}
	\label{fig:7}
\end{figure}

As evidenced in Figure~\ref{fig:5}, dimensions 1, 9, 11, 13, 24, and 28 exhibit consistent negative correlations with both betweenness and reciprocity—the greater their absolute values, the stronger the adverse effects. For non-neuronal, dimension 11 demonstrates a monotonically increasing influence, as does 28, while dimensions 9 and 1 show monotonically decreasing trends. Dimension 24 displays nonlinear behavior: when $z_{24} > 0$, it initially exerts positive influence that gradually diminishes into negative effects, whereas $z_{24} < 0$ produces strictly negative impacts that intensify with decreasing values. Dimension 13 presents the inverse pattern of 24. Other dimensions show negligible effects. These effects can be verified in Figure~\ref{fig:6}.

As shown in Figure~\ref{fig:7}, despite these identifiable patterns, dynamic programming yields suboptimal results for these three features. We attribute this to inherent feature complexity—the strong coupling between dimensions induces excessive variance in SHAP values, thereby destabilizing the dynamic programming optimization.

\section{Reconstruction Performance}
\label{app:recon}

\begin{table}[h]
\centering

\begin{threeparttable}
\caption{Reconstruction performanc on FlyWire. We evaluate (1) Edge reconstruction performance using AUC and accuracy between original and reconstructed adjacency matrices, and (2) Node category recovery using classification accuracy and F1-score comparing original versus predicted node categories.}
\label{table:2}
\begin{tabular}{lcccc}
\toprule
\multirow{2}{*}{\textbf{Method}} & 
\multicolumn{4}{c}{\textbf{FlyWire}}  \\
\cmidrule(lr){2-5} 
& \textbf{Node Acc.$\uparrow$} & \textbf{Node F1$\uparrow$} & \textbf{Edge Acc.$\uparrow$}  & \textbf{Edge AUC$\uparrow$}\\
\midrule
Ours &  0.982 & 0.979 &  0.978 & 0.959\\
\bottomrule
\end{tabular}
\end{threeparttable}
\end{table}

The reconstruction results in Table~\ref{table:2} demonstrate our model's ability to reasonably recover both node categories and edge connectivity, though with some expected information loss. We achieve 0.982 accuracy in node classification and 0.978 edge reconstruction accuracy, with slightly lower performance in edge AUC (0.959), suggesting the model captures major structural patterns while missing some finer connection details. These metrics reflect the inherent trade-offs of our approach - while the information bottleneck causes some reconstruction imperfection, it enables the low-dimensional analysis that is central to our framework. The performance is encouraging given this constraint, though we recognize these results would benefit from comparison with standard reconstruction baselines in future work. The modest differences between node and edge metrics may represent the necessary compromise between reconstruction fidelity and analytical tractability that defines our method's design philosophy.

\begin{figure}[!htbp]
\centering
\includegraphics[width=\textwidth]{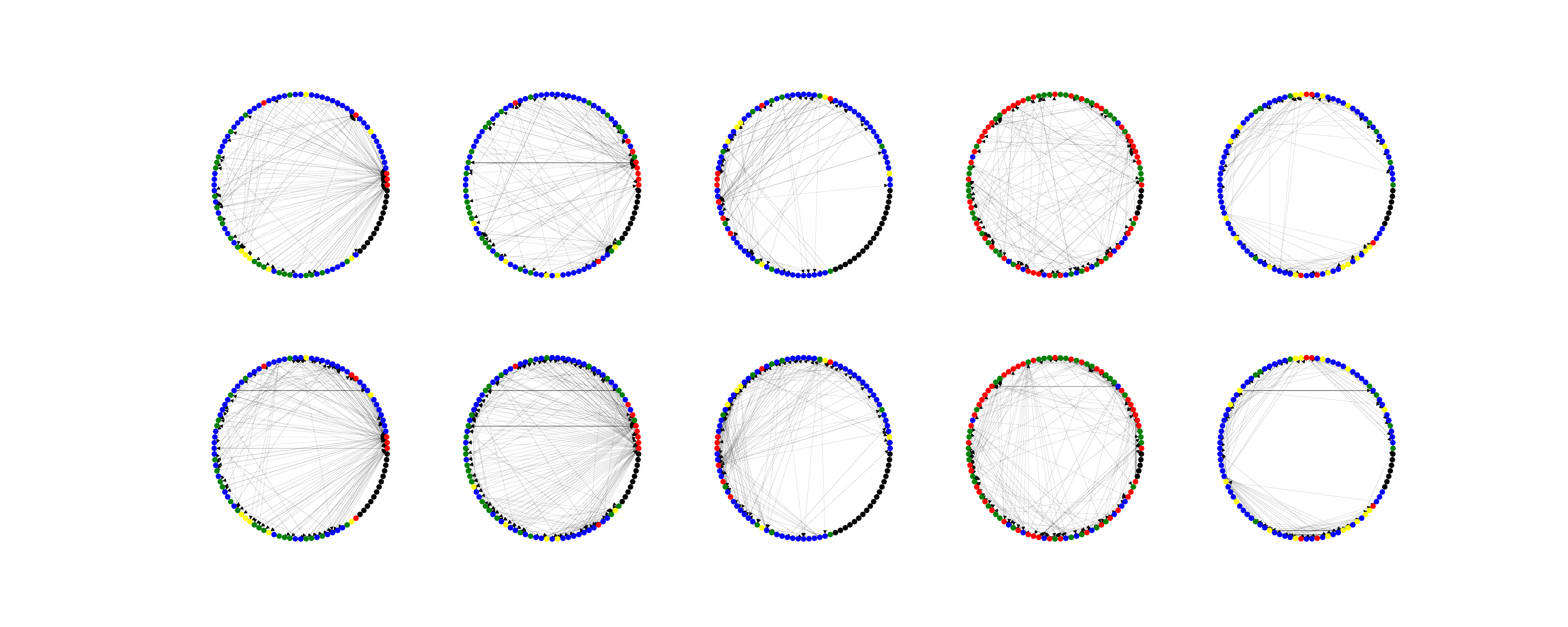}
\caption{Graph reconstruction results: Original samples (top) and their reconstructed counterparts (bottom), with node colors representing type categories.}
\label{fig:8}
\end{figure}

Figure~\ref{fig:2} demonstrates our model's graph reconstruction capability by comparing original connectome samples (top row) with their reconstructed counterparts (bottom row). The visual comparison shows that the model successfully preserves the overall topological structure and node type distributions (represented by consistent color patterns), while some subtle differences in local connectivity can be observed upon closer inspection. The reconstructions maintain the characteristic clustering patterns of neural circuits, though with minor variations in edge density and connection specificity that reflect the expected information loss through our model's bottleneck architecture. These results align quantitatively with the reconstruction metrics reported in Table~\ref{table:2}, providing visual confirmation that while the model captures essential connectome features, perfect reconstruction remains challenging due to our framework's emphasis on dimensional reduction and interpretability over absolute fidelity. The preserved node category assignments (colors) are particularly noteworthy, suggesting the model reliably maintains neurotransmitter-type information during encoding-decoding.

\end{document}